# Multimodal and multicontrast image fusion via deep generative models

Giovanna Maria Dimitri (1,5)*, Simeon Spasov (1)*, Andrea Duggento (2), Luca Passamonti (3,6), Pietro Lió (1), Nicola Toschi (2,4)

1) University of Cambridge, Cambridge, Department of Computer Science and Technology, William Gates Building, 15 JJ Thomson Ave, Cambridge, CB3 0FD, UK

2) Department of Biomedicine and Prevention, University of Rome "Tor Vergata", Via Montpellier 1, 00133, Roma, RM, Italy

3) Department of Clinical Neurosciences, University of Cambridge, Herschel Smith Building, Forvie Site, Robinson Way, Cambridge Biomedical Campus, Cambridge, CB2 0SZ, Cambridge, UK

4) A.A. Martinos Center for Biomedical Imaging, Massachusetts General Hospital and Harvard Medical School, Boston, USA

5) Dipartimento di Ingegneria dell'Informazione e Scienze Matematiche (DIISM), University of Siena, Italy

6) Istituto di Bioimmagini e Fisiologia Molecolare (IBFM), Consiglio Nazionale delle Ricerche (CNR) Via F.lli Cervi, 93, 20090, Segrate, Milano, Italy.

*The two authors contributed equally to this work

**Corresponding Authors:**

-Giovanna Maria Dimitri
**Address:** Department of Computer Science and Technology, University of Cambridge,
William Gates Building, 15 JJ Thomson Ave, Cambridge, CB3 0FD,
United Kingdom
**Email**: gmd43@cl.cam.ac.uk, giovanna.maria.dimitri@gmail.com

-Simeon Spasov
**Address:**
Department of Computer Science and Technology, University of Cambridge,
William Gates Building, 15 JJ Thomson Ave, Cambridge, CB3 0FD,
United Kingdom
**Email**: simeon.spsv@gmail.com

# Abstract

Recently, it has become progressively more evident that classic diagnostic labels are unable to accurately and reliably describe the complexity and variability of several clinical phenotypes. This is particularly true for a broad range of neuropsychiatric illnesses such as depression and anxiety disorders or behavioural phenotypes such as aggression and antisocial personality. Patient heterogeneity can be better described and conceptualized by grouping individuals into novel categories, which are based on empirically-derived sections of intersecting *continua* that span both across and beyond traditional categorical borders. In this context, neuroimaging data (i.e. the set of images which result from functional/metabolic (e.g. functional magnetic resonance imaging, functional near-infrared spectroscopy, or positron emission tomography) and structural (e.g. computed tomography, T1-, T2- PD- or diffusion weighted magnetic resonance imaging) carry a wealth of spatiotemporally resolved information about each patient's brain. However, they are usually heavily collapsed *a priori* through procedures which are not learned as part of model training, and consequently not optimized for the downstream prediction task. This is due to the fact that every individual participant usually comes with multiple whole-brain 3D imaging modalities often accompanied by a deep genotypic and phenotypic characterization, hence posing formidable computational challenges.

In this paper we design and validate a deep learning architecture based on generative models rooted in a modular approach and separable convolutional blocks (which result in a 20-fold decrease in parameter utilization) in order to a) fuse multiple 3D neuroimaging modalities on a voxel-wise level, b) efficiently convert them into informative latent embeddings through heavy dimensionality reduction, c) maintain good generalizability and minimal information loss. As proof of concept, we test our architecture on the well characterized Human Connectome Project database (n=974 healthy subjects), demonstrating that our latent embeddings can be clustered into easily separable subject strata which, in turn, map to different phenotypical information (including organic, neuropsychological, personality variables) which was not included in the embedding creation process.

The ability to extract meaningful and separable phenotypic information from brain images alone can aid in creating multi-dimensional biomarkers able to chart spatio-temporal trajectories which may correspond to different pathophysiological mechanisms unidentifiable to traditional data analysis approaches. In turn, this may be of aid in predicting disease evolution as well as drug response, hence supporting mechanistic disease understanding and also empowering clinical trials.

**Keywords**

- Deep Autoencoder
- Phenotype Stratification
- Latent Embeddings
- Precision Medicine
- Separable Convolutions
- Multimodal Neuroimaging

# 1. Introduction

Over the last years, it is has become progressively more evident that classic diagnostic labels are unable to accurately and reliably describe the complexity and variability of several clinical phenotypes. This is particularly true for a broad range of neuropsychiatric illnesses such as depression and anxiety disorders or behavioural phenotypes such as aggression and antisocial personality. In neurology and psychiatry, the problem is compounded by the clear lack of boundaries between normal and abnormal behaviour and by the presence of often overlapping diagnostic categories. The need for a trans-diagnostic psychiatry and for a brain-based categorization of the clinical entities that transcend the traditional Diagnostic Standard Manual (DSM) labels has been emphasized several times. Alzheimer's Disease (AD) and Parkinson's Disease (PD) are paradigmatic examples. Both diseases manifest themselves with a myriad of symptoms and signs which may or may not occur across different patients, and typically can be linked to heterogeneous timescales and trajectories (i.e. evolutions over time). Accordingly, *post-mortem* studies have confirmed an high heterogeneity in neuropathological findings not only in AD (Rabinovici et al., 2017) but also in other, classically unrelated, but most probably somehow dimensionally connected neurological syndromes (Adler et al, 2010).

To overcome this issue, trans-diagnostic approaches that rely on a more nuanced and mechanistic representation of the demographic and clinical features of each individual have been proposed. More specifically, patient heterogeneity can be better described and conceptualized by grouping individuals into novel categories, which are based on empirically-derived sections of intersecting *continua* that span both across and beyond traditional categorical borders. Such novel patient groupings are likely to be disjoint and possibly become independent from the currently employed disease categories. This is because a single patient condition often transcends the traditional taxonomies. In contrast, it is reasonable to visualize and describe each patient as occupying their own, unique position in a high-dimensional space that depends on specific pathophysiological mechanisms. The uniqueness of this position in such a high-dimensional space is likely to translate into new opportunities to optimize the diagnosis and treatment to the individual needs of each patient ('personalized medicine'). However, the possibility to translate this into clinical practice can only depend on the possibility to access the multidimensional biomarker space that defines the uniqueness of each patient's neurocognitive and behavioural "profile".

This conceptual paradigm is often termed precision- and (in some cases) personalized medicine, which lies in stark contrast with the more commonly employed "one size fits all" approach. The latter is still pervasive in most clinical disciplines, although exceptions exist like e.g. in oncology, where some degree of personalization has been clinically successful. The root of this discrepancy lies largely in the difficulty to handle/cluster very high dimensional data (like e.g. multimodal imaging) with high spatial correlations, presenting a major knowledge gap for the end-user (e.g. clinicians and neuroscientists). In this sense, having a low dimensional representation of high-dimensional data could be instrumental in e.g. identifying disease subtypes based on objective patient data as well as associations of such disease subtypes with differential responses to therapy and or different clinical outcomes. In turn, this would empower clinical trials by uncovering possible patient trajectories. Such observations highlight the urgent need for more objective classification criteria and frameworks, which need to be based on measurable and reproducible biomarkers. This can be achieved by employing deep generative models that are able to meaningfully group individuals (healthy or with clinical disorders) on the basis of key brain structural measures.

Neuroimaging data carry a wealth of spatiotemporally resolved information about each patient's brain. However, they are usually heavily collapsed *a priori* through procedures which are not learned as part of model training, and consequently not optimized for the downstream prediction task. Still, for a comprehensive data-driven stratification, all relevant pathophysiological mechanisms should be well-represented in the multidimensional data fed into the aggregation and pattern recognition frameworks. In this context, the recent appearance of several large multicentre and multimodal, curated large (between 1000 and 40000 individuals at the time of writing) data repositories (e.g. the Parkinson Progression Marker Initiative (PPMI) (Jennings et al., 2011), the Alzheimer Disease Neuroimaging Initiative (ADNI) (Mueller et al., 2005), the UK Biobank initiative (Sudlow et al., 2015), the Cam-CAN dataset (Taylor et al 2017) and the Human Connectome Project (HCP) (Van Essen et al., 2013) is providing novel formidable opportunities as well as challenges. On one hand, this amount of information which has never previously been accessible to neuroimaging researchers: every individual participant usually comes with multiple whole-brain imaging modalities of $10^5$-$10^6$ voxels each, often acquired at multiple timepoints. Importantly, these data are usually accompanied by a deep geno/phenotypic characterization (e.g. genetic, biochemical, bio humoral, and neuropsychological markers). This opens up avenues to robust cross-modality data fusion, and therefore to subsequent reduction into embeddings that are fine-grained enough to 1) inform mechanistic hypotheses about disease physiology as well as 2) about the neural substrates determining currently unexplainable within-disease variability. Computational and conceptual challenges in designing data reduction architectures able to exploit voxel-wise multimodal 3D imaging data (most deep learning frameworks are designed to learn from 2D images) while retaining realistic computation times and, crucially, extract informative embeddings while reducing data dimensionality by at least a factor 1000, are severe.

The aim of this paper is therefore to design and validate a deep learning (DL) architecture based on generative models rooted in a modular approach and separable convolutional blocks. Our goal is to efficiently extract low level informative embeddings while A) fusing multiple 3D neuroimaging modalities on a voxel-wise level B) performing heavy dimensionality reduction with minimal information loss and C) building an architecture able to efficiently decode brain images. This latter aspect is in fact gaining much attention in the field of DL and neuroimaging. A variety of generative approaches have been proposed for image restoration (e.g. denoising [Bermudez et al. 2018, Benou et al., 2018, P. Vincent et al., 2010], artefacts removal [S. Liu et al., 2021] or reconstruction of subsampled MR images (Falvo et al 2021, Falvo et al. 2019 ), image super-resolution (A.S. Chaudhari et al. 2018), synthesis (H. C. Shin et al., 2018, D. Nie et al., 2017, M. Frid-Adar et al., 2018, Yurt et al 2021) registration (G. Balakrishnan et al., 2019, B.D. de Vos et al., 2019) and segmentation (Bernal et al., 2018). All these methods rely on an autoencoding or adversarial training setup (see also Yang et al 2020, Ma et al 2020 and refer to Lundervold et al., 2019 for a thorough review). More recently, methods for dealing with data paucity in medical imaging have also been pioneered, e.g. transfer learning (Dar et al., 2020 a.), parameter-efficient architectures (Dar et al., 2019), prior-guided learning (Dar et al., 2020 b.), as well as approaches which combine information from multiple modalities (Yurt et al. 2021 and Xu et al. 2021). Moreover, deep learning methods have been shown to improve reconstruction results in multicontrast images (Liu et al 2021, Do et al 2020).

As proof of concept, we test our architecture on the well characterized Human Connectome Project (HCP) dataset, where we used multiple modalities to perform our experiments such as JAC, NDI, FA and T1-W brain scans. We demonstrate that our latent multimodal embeddings can be clustered into easily separable subject strata. Additionally, these strata map to different phenotypical information (including organic, neuropsychological, personality variables) which was not included in the embedding creation process.

## 2. Materials and Methods

A comprehensive overview of the workflow adopted in this paper is presented in Fig. 1.

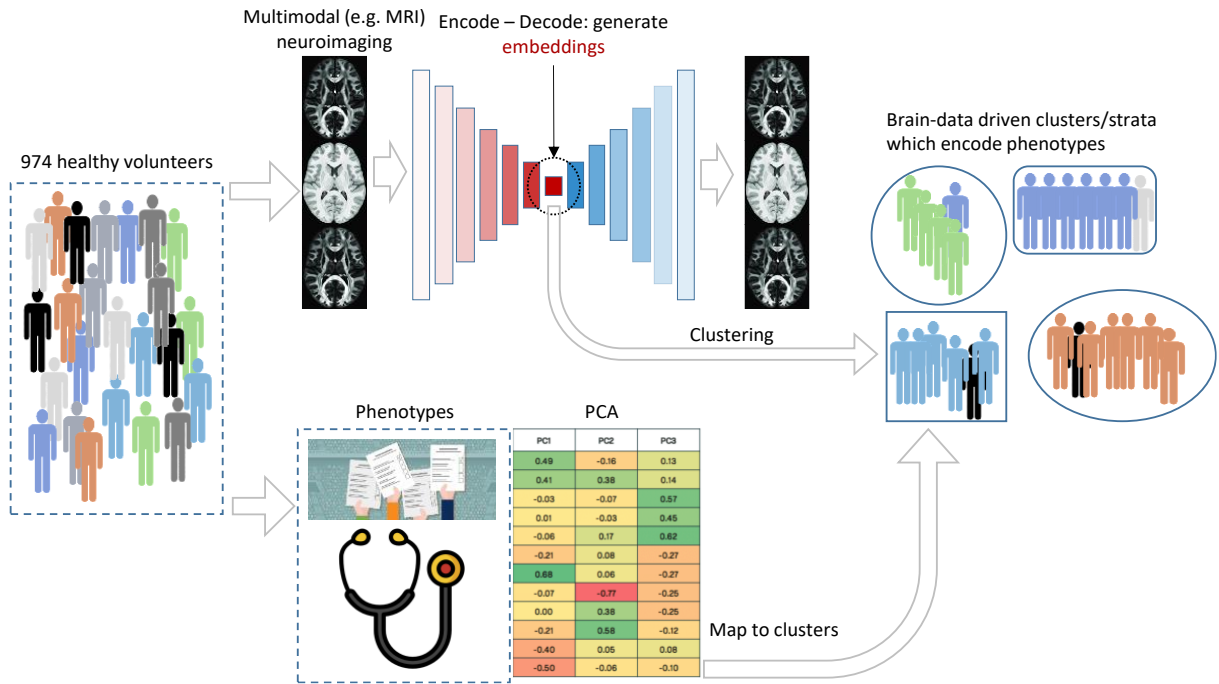

*Figure 1: A summary of the workflow of our paper. First, we design an efficient multi-modal deep learning architecture to encode subjects in low dimensional embeddings. Our model is unsupervised, which means that it is trained to decode the original inputs. The learnt embeddings summarize information from multiple modalities. Then, the model upsamples the embeddings to generate a separate decoding for each input modality, using information from both modalities. Second, we cluster the subject embeddings using affinity propagation to stratify the subject population by identifying separate subtypes/subgroups. Third, we aim to demonstrate the external validity to our framework by mapping each of the identified clusters with external phenotypic traits/factors (which have been generated from a number of raw phenotypic scores and interpreted by an expert neuroscientist, LP) which did not concur to forming the embeddings. Loadings scores of the PCA transformation (for the first three Principal Components extracted) are shown in an exemplary manner in the table above in the figure. This is done by simply averaging the factor loadings within each cluster for each factor. This results in clustering the population using only brain imaging data. Clusters are then mapped to each of the factors.*

## 2.1 Multi-Modal Deep Learning architecture

*2.1.1 Overview*

The architecture designed for this study can be viewed as a deep learning, unsupervised autoencoder, composed of a downsampling stage (encoder) and an upsampling stage (decoder) (See Fig 2). Several details are inspired from our recent supervised dual learning framework designed to process 3D medical images (Spasov et al., 2019). Here, on top of adopting an encode-decode design, in order to improve granularity, we incorporate skip connections between operational blocks at both stages (Ronneberger, et al 2015). The key aspects of our design choices can be summarized as: A) Parameter efficiency: in addition to conventional convolutional layers, we employed alternating sequences of depth-wise and pointwise convolutions, also known as depth-wise separable convolutions (F. Chollet 2017, A.G. Howard et al. 2017). This results in a lightweight neural network architecture with high parameter efficiency. This strategy has achieved superior performance in classification as well as image segmentation tasks (F. Chollet 2017, A.G. Howard et al. 2017, Zoph et al. 2018) and has been extended to medical image classification (Spasov et al. 2019), while limiting overfitting. B) Multi-modality: it is well established that combining various medical imaging modalities can greatly enhance diagnostic and prognostic performance. We incorporate this data fusion aspect by employing a separate encoder and decoder for each of the modalities, which then all concur to generate a common embedding. The embedded representation is therefore learned (and contains information) from all modalities and can also be used for downstream learning tasks (see Section 3.2. and thereafter for a proof-of-concept experiment on multimodal neuroimaging data). Crucially, the decode step is based on this multimodal embedding only, demonstrating that modality-specific information can be re-distilled from the fused, embedded product of the encoding branch. C) U-Net skip connections: The encoders in our architecture reduce image dimensionality by consecutively pooling the output features of previous layers. This loss of resolution poses a difficulty when learning to decode high-fidelity images with good granularity. U-Net (Ronneberger, et al 2015) overcomes this challenge by supplementing the encoder-decoder architecture with contracting paths (or skip connections) between layers in the downsampling and layers in the upsampling stages. In this way we transfer localized high-resolution information directly to the decoding phase. In accordance with the original U-Net paper, we double the number of feature channels after a pooling operation of size and stride 2 in each image dimension during downsampling, and half the number of feature channels during upsampling. Please note that in this paper, the wording "single channel" refers to the fact that most 3D medical images are greyscale (i.e. one channel) as compare to e.g. natural images in RGB (i.e. 3 channels).

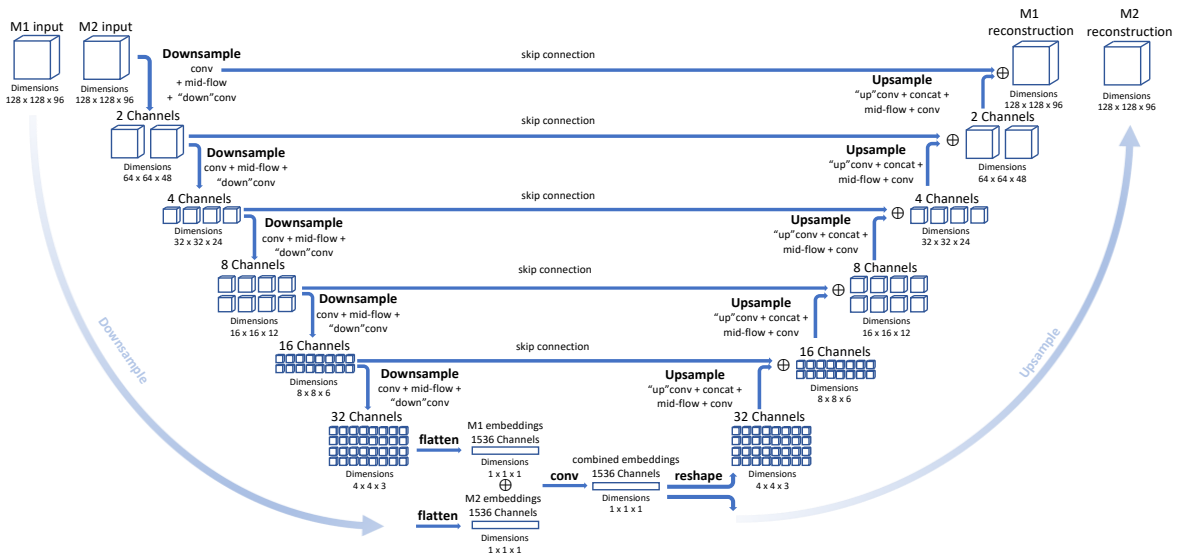

*Figure 2: Our multi-modal architecture for unsupervised learning. The network simultaneously takes multiple modalities (two in this diagram, M1 and M2, however the architecture is able to accept an arbitrary number of modalities) and decodes both original inputs from a jointly learnt low dimensional embedding, denoted as "combined embeddings" in the figure. All modalities are processed in the downsampling and upsampling stages by successive applications of ad-hoc operational blocks termed conv, mid-flow, "down"conv and "up"conv. We reduce the dimensionality of the inputs via downsampling, and we merge their representations via concatenation ($\oplus$ symbol) and convolution. Then, each upsampling stage can be viewed as a mirror image of a downsampling stage at a certain depth. We concatenate corresponding feature maps between the upsampling and downsampling stages, and apply a series of "up"conv, mid-flow and conv block operations to recover the original dimensionality of the inputs. Network parameters are learnt by minimizing the binary cross-entropy between the decoded and input images.*

### 2.1.2 Architecture details

Figure 2 depicts the network architecture of our deep learning autoencoder for medical image embedding and decoding (two modalities in the diagram). All modalities have a single channel and voxel values are scaled in the [0, 1] range prior to processing. In this paper, all training was performed with two modalities (M1 and M2, see Section 2.3 for data details, possible combinations and Section 3. for realistic examples), where each input was a 3D image of size 128x128x96, resulting in approximately 12.2 M data points per modality. The joint multimodal embedding vector is composed of 1536 elements (hence representing an approximately 16000-fold data reduction in the encode step) end employed to recover the initial dimensionality, contrast and information of the inputs. Both the encode and decode stages are implemented as a sequence of operational blocks termed conv, mid-flow,

"down"conv and "up"conv (see below). The parameter-efficient separable convolutions (Spasov et al 2019) are integrated within the mid-flow block and comprise the majority of convolutional procedures in the network. At each downsampling step we apply a sequence of operations comprising "conv", mid-flow and "down"conv, and during upsampling – "up"conv, concat (concatenation needed for skip connections), mid-flow and conv. The inner working of these operational blocks is depicted in Fig. 3. In the downsampling stage we consecutively reduce the dimensionality of the inputs in 5 steps. At each step each dimension of the 3D images is halved, and the number of convolutional filters is doubled. We undo this process during upsampling by upscaling the embeddings by the same factor of 2 at each step. The downsampling and upsampling streams process all modalities in parallel. We only combine the low-dimensional representations of each input by first flattening them separately, then concatenating (denoted by ⊕ symbol in Fig. 2) and convolving them. To illustrate how high granularity is achieved through skip connections, consider the output feature maps with dimensions 32x32x24 and 4 channels in the upsampling stage. First, we "up"conv these activations and scale each dimension twice to 64x64x48. Then, we concatenate these intermediate representations with feature maps produced after a single step of downsampling, which have the same dimensionality, as denoted by the directed arrow labelled "skip connection" in Fig. 2. We then mix the combined representations with a mid-flow and a conv block to produce the upsampled feature maps one level higher in the Upsample stream (dimensions 64x64x48 with 2 channels).

*2.1.3 Operational blocks*

In both the encoding and decoding stages, we employ a series of operational blocks, similar to the ones proposed in Spasov et al., 2019, comprising conv, "up"conv, "down"conv, mid-flow (see Fig. 2 for a schematic scheme of the whole DL network). Fig.3 depicts the most foundational block in our neural network model – the (sep)conv operational block. It has two variants – one working with a standard 3D convolutional operator (*conv*) and another utilizing 3D separable convolutions (*sep conv*). In both variants the inputs are first convolved, then batch normalization (Ioffe and Szegedy 2015) and an exponential linear unit activation are sequentially applied. We also allow for dropout (Srivastava, et al 2014) to be used as the last layer in the (sep)conv block. Both "down"conv and "up"conv are simple extensions of the conv block. For "down"conv we append 3D Max Pooling (size=3, stride=1) to reduce the dimensionality of feature maps in the downsampling stream after conv block processing. On the other hand, in "up"conv, we prepend 3D upsampling (size doubles in each dimension) to the conv block. The majority of convolutional processing occurs in an operational block termed mid-flow (see Fig 2. b), which has two parallel lines of processing. On one hand, three consecutive sep convs are applied to the inputs. On the other, we introduce a skip connection which does not transform the inputs and allows them to propagate unchanged to the output, which has been proposed to facilitate training of deeper neural networks (He et al., 2016). Finally, we combine the parallel lines of processing by adding them and returning a single output feature map. Substituting standard convolutions with separable ones results in a 20-fold decrease in parameter utilization in a single mid-flow block. A thorough presentation of separable convolutions and their functional difference from standard convolutions is given in F. Chollet, 2017 and their implication to parameter efficiency in medical imaging application is discussed in Spasov et al., 2019.

The model was implemented in Keras, with a TensorFlow backend and trained on a Nvidia TITAN V GPU. The training was performed over 200 epochs, batch size 1. We set the dropout rate at 0.1 for all conv and sep conv operational blocks and apply an L2 regularization coefficient of $5\times10^{-5}$ for all network parameters. We use the Adam optimizer (Kingma et al., 2014) at its default settings to train the neural network.

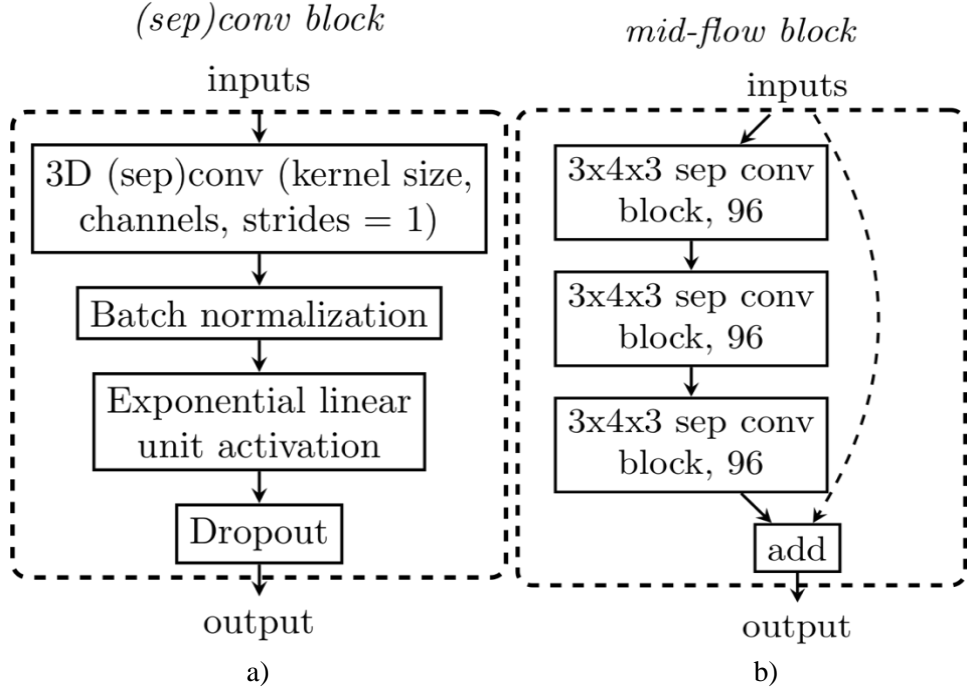

*Figure 3: Inner workings of the fundamental blocks used in the neural network diagram from Fig. 2. Since the conv, sep conv, "down" conv and "up" conv blocks are very similar in structure, we present a single diagram in fig 2. a) and use it to describe the other derived operational blocks. For example, in "down" conv we apply 3D Max Pooling (size=3, stride=1) at the end of processing, whereas in "up" conv, we prepend 3D Upsampling (size doubles in each dimension); sep conv denotes using separable instead of conventional convolutions. Fig 1. b) presents the mid-flow block based on a series of three sep conv blocks and a skip connection adding the original input and the output of the final separable convolution. All convolutional kernel sizes in our work are of size 5x6x5 with stride=1. The number of channels can be inferred from Fig. 2.*

**2.2 Cross-Validation Procedure and Assessment of the decoding quality**

In order to evaluate the performance of our encoder-decoder architecture, and verify our model did not suffer from overfitting, we implemented a 10-fold cross validation procedure using two modalities from our data set of approximately 1000 subjects. The following procedures / metrics were employed for each fold to assess the "difference" between the original and decoded images, and hence the overall decoding performance.

1) Mean Squared Error (MSE) computation. The MSE between each decoded and original pair was computed across whole voxelwise images.

2) Definition of a measure called henceforth "normalized difference" (NormDiff), defined as $NormDiff = \frac{reconstructed - real}{reconstructed + real}$. This metric is bounded between -1 and 1, and is intended to quantify asymmetry between a real and a decoded value. The advantage of this is to eliminate dependency on the underlying image intensity and to focus on relative errors. By calculating this metric voxel-wise, we obtained normalized difference images which are then collapsed through the median operator to obtain one value for each real-decoded pair.

3) Estimation of Contrast to noise ratio (CNR), an important metric in diagnostic imaging. We adopted a simple definition of CNR based on two randomly selected regions of interest (ROI), namely $CNR_{1,2} = \frac{mean(ROI1) - mean(ROI2)}{std(ROI1 + ROI2)}$. We then divided each image in regular regions of interest (ROIs) of dimension 4x4x3, and successively randomly sampled 1000 ROI pairs while avoiding sampling from the background. This yielded a 1000-sample estimate of the CNR distribution of each image, which was collapsed through the median operator to obtain one value for each real image and one value for each corresponding decoded image. Finally, the decoded-real difference in estimated median CNR was evaluated using the normalized difference as defined above.

In addition, in order to demonstrate the ability of our architecture to include multiple modalities while still retaining robustness in the decoding of any particular modality, we ran experiments where every modality was decoded using a combination of 1, 2, 3, or 4 modalities while quantifying both MSE and CNR differences. These experiments were run in a randomized 70/30 train/test setup for computational efficiency.

### 2.3 Data and pre-processing

*2.3.1. Population*

The population used for the present study consists in 974 subjects drawn from the publicly available Human Connectome Project (HCP) public data repository located at https://humanconnectome.org/ (see Table 1 for demographics). The cohort resulted from down selecting the total 1200 subject to those who had valid and downloadable diffusion MRI data (see Table 1) at the time of processing. Additional phenotypic variables are summarized in the supplementary materials (Table S1).

|  | *Age* | *Handedness* | *BMI* |
|---|---|---|---|
| *Males (n=444)* | 27.94±3.65 | 60.54±43.63 | 26.78±4.36 |
| *Females (n=530)* | 29.42±3.52 | 69.88±43.93 | 26.10±5.85 |

*Table 1: The table shows median and standard deviation of three demographics information on the HCP cohort studied: age, handedness and BMI (n=974)*

*2.2.2. Phenotypic Variables and factor analysis*

In order to better characterize our study population, and also in order to test the ability of the information encoded in our multimodal embeddings to map to distinct phenotypic groups, we clustered the population in the latent embeddings space. For this purpose, 51 questionnaire measures, demographic variables, or neuropsychological tests assessing for lifestyle, psychological and physical well-being, personality, cognitive functioning, and emotional behavior were included in this study (see Supplementary material). A principal component analysis (PCA) with Varimax rotation was used to reduce data dimensionality and identify the main factors for the psychological, cognitive, and physical and mental health latent dimensions. Varimax rotation rotates the orthogonal basis to dispersion of the loading scores across components, hence simplifying the expression of the subspace extracted. The rotated component matrix was employed for extraction of individual factor (retaining only factors which corresponded to eigenvalues > 1) loading scores using the regression methods. Those scores were then employed for characterizing the strata generated by unsupervised clustering of our multimodal embeddings (see Section 2.4). In details, after component extraction, the values of the factors for each individual are commonly extracted using simple regression of the original values against the components. In order to provide an operational neurobiological interpretation for each factor, ladings were thresholded at an absolute value of 0.3, which allowed an expert neuroscientist (LP) to identify the underlying factors and interpret them according to common associations between phenotypic variables.

*2.3.3 MR data acquisition*

All imaging data employed in this study were acquired by the HCP consortium on a Siemens Skyra 3T scanner with a customized SC72 gradient insert. T1w 3D MPRAGE images were acquired with TR=2400 ms, TE=2.14 ms, TI=1000 ms, flip angle=8 deg, FOV=224x224, 0.7 mm isotropic voxel, bandwidth =210 Hz/px, iPAT=2, Acquisition time=7:40 (min:sec). Diffusion weighted images were acquired with Spin-echo EPI sequences (b-values = 0, 1000, 2000, 3000 s/mm2 in 90 gradient directions (interspersed with an approximately equal number of acquisitions on each shell). Diffusion weighting consisted of 3 shells of b=1000, 2000, and 3000 s/mm2. The diffusion directions were uniformly distributed in multiple q-space shells and optimized so that every subset of the first M directions is also isotropic (Caruver et al., 2013), TR=5520 ms, TE=89.5 ms, flip angle=78 deg, refocusing flip angle=160 deg, FOV=210x180 (RO x PE) matrix=168x144 (RO x PE), slice thickness =1.25 mm, 111 slices, 1.25 mm isotropic voxels, Multiband factor=3, Echo spacing=0.78 ms, BW=1488 Hz/Px, Phase partial Fourier 6/8). A full diffusion MRI session included 6 runs (approximately 9 minutes and 50 seconds each). Diffusion gradients were monopolar. Image reconstruction uses SENSE multi-channel (Sotiropoulos et al., 2013).
See https://humanconnectome.org/storage/app/media/documentation/s1200/HCP_S1200_Release_Reference_Manual.pdf. for additional details. All brain images were reviewed for incidental brain abnormalities by a neuro-radiologist,and pre-processed following the minimal HCP pipeline by the HCP consortium (Glasser et al., 2013).

*2.3.4 Diffusion data pre-processing and model fitting*

Diffusion image preprocessing, performed by the HCP consortium, included state-of the art procedures: intensity normalization across runs, distortion correction through the 'TOPUP' tool (part of FSL, (Jenkinson et al., 2012), eddy current and motion correction through the 'EDDY' tool (also part of FSL), gradient nonlinearity correction, calculation of resulting gradient bvalue/bvector deviation, and Registration of mean b0 to the corresponding T1w volume with FLIRT BBR+bbregister (also part of FSL). This is followed by transformation of diffusion data, gradient deviation, and gradient directions to 1.25mm structural (T1w) space. Starting from these preprocessed data both the tensor model (Diffusion Tensor Imaging – DTI, only bvalues <= 1000) and the NODDI model (all data) (Zhang et al., 2012) were fitted to each individual dataset using the microstructure diffusion toolbox (MDT) (https://github.com/robbert-harms/MDT). From the DTI model, we extracted fractional anisotropy maps (FA), which are known to be sensitive to microstructural alterations. From the NODDI model we extracted neurite dispersion indices maps (NDI), which are known to be even more specific to the same type of alterations.

*2.3.5 Study specific template creation and image registration*

To improve registration accuracy, we built a study-specific T1 template using the T1-weighted images After bias filed correction in FSL, all T1-weighted images were non-linearly co-registered to each other (symmetrical diffeomorphic mapping), averaged iteratively (5 iterations), and co-registered to the final template. A binary brain mask was generated from the T1 template using BET2, also part of FSL.
The template creation and registration procedures were performed using the Advanced Normalization Tools (ANTs) software (Avants et al., 2011). Non-linear transformations were initialized through a chain of center of mass alignment, rigid, similarity, and fully affine transformations followed by local nonlinear warping (metric: neighborhood cross correlation, sampling: regular, gradient step size: 0.12, four multi-resolution levels, smoothing sigmas: 3, 2, 1, 0 voxels in the reference image space, shrink factors: 6, 4, 2, 1 voxels, histogram matching, data winsorisation with quantiles: 0.001, 0.999, convergence when the slope of the normalized energy profile over the last 10 iterations<$10^{-8}$).
We then extracted the local Jacobian determinant (JAC) of the non-linear part of the last-stage deformation field which takes each T1-image into template space. The JD image quantifies the amount of local volume variation (contraction/expansion) computed when matching the single subject image to the template and is generated in the template space. Also, the same warp fields were applied to all diffusion derived maps (FA and NDI), resulting in all images (T1, FA, NDI, JAC) in the same 1.25 mm resolution space with high intra-modality and inter-subject anatomical correspondence and downsampled by a factor 2 to yield final volumes of size 128x128x96 (Figure 3), which match common resolutions employed in clinical and neuroscience applications. Finally, all images were robustly normalized into the 0-1 range by using min-max normalization (0.1 and 99.9 percentile) across the whole population. Min-max normalization is often employed in deep learning to facilitate convergence of the learning algorithm. Image co-registration required ~26,500 hours of CPU time. Calculations were performed on a 600-node compute cluster with 8GB-RAM/node. It is important to note that while the registration is instrumental to create the JD maps as well as achieving precise anatomical correspondence, it has been shown that the key elements taken from our architecture perform just as well using natural (i.e. unregistered) images (Ramon-Julvez et al., 2020).

**2.4 Mapping multimodal embeddings to phenotypic variables**

*2.4.1 Clustering multimodal embeddings*

As proof of concept of the usefulness of our multimodal embedding in efficiently compressing information, we adopted an unsupervised clustering strategy in order to explore the existence of subgroups in the brain data (a near-impossible procedure when working directly with whole brain images) and mapped these subgroups onto the external phenotypic data which had not concurred to forming the embeddings. We employed affinity propagation (AP) clustering, which does not require an a priori number of clusters to be specified and performs well in the presence of noise (Vlasblom et al 2009). AP was used in conjunction with Euclidean Distances and relies on two main parameters: preference (which determines the likelihood of a point to be chosen as exemplar) and damping (related to the speed and accuracy of numerical stabilization). These were optimized through a grid search procedure across 10 damping (0.5-1, equal intervals) and 1000 preference values (sampled with equal spacing between bottom and top percentile of the negative squared Euclidean Distances of the data matrix at hand) using the Silhouette index score as a performance metric (Rousseeuw et al 1987).
The between-cluster differences in factors extracted from the phenotypic variables were assessed through the non-parametric Kruskall-Wallis tests (KW) (Wayne, 1990) and corrected for multiple comparisons across factors through the FDR Benjamini and Hockberg correction procedure (Benjamini & Hochberg, 1995). P<0.05 (FDR corrected) was considered statistically significant.

*2.4.2 Bootstrap analysis between-cluster differences in phenotypical factors*
In order to test the hypothesis that the statistically significantly differences in phenotypic variables we found in the previous step may be a result of random effects, we proceeded as follows. Given $m$ subjects (i.e. embeddings) which have been partitioned into $n$ "true" clusters of sizes $s_1...s_n$, as in 2.4.1, we randomly repartition the $m$ subjects in the clusters into same number of groups ($n$) with the same sizes ($s_1...s_n$) as the "true" clusters and compute the corresponding KW statistics. This bootstrapping procedure is repeated M times (here M = 10000, sampling with replacement). We then evaluated the p-value related to the hypothesis described above as the number of bootstraps whose KW statistics is higher than the "true" KW statistics, divided by M. This fraction of bootstraps serves as a surrogate p-value corresponding to the above outlined hypothesis – i.e. if this pvalue is <0.05 (after FDR correction across all factors), the "true" statistical significance in clusterwise differences between factor values can be considered non-random.

# 3 Results

## 3.1. Decoding results

In order to evaluate the quality and generalization ability (i.e. the model's ability to perform well on an unseen test set) of our model, we performed detailed experiments using three pairs of modalities i.e. JAC-FA, JAC-NDI, T1-NDI (Figure 4). These choices of pairs were made in order to a) be possibly sufficiently general to explore the ability of our architecture to combine different contrasts, and 2) generate pairs of modalities which, to some degree, complementarily included mainly white matter (WM) and grey matter (GM) contrast (i.e. imaging modalities able to mainly highlight and outline WM and GM, respectively). The presence, in the full decoded images, of key anatomical detail was confirmed through sample whole-image inspection by an expert neurologist (LP). Visually, we obtained good decoding quality in FA, NDI and JAC images (Figure 4). Some loss of detail in the subcortical structures as imaged in the T1 contrast can be seen when T1 is paired with NDI. This is possibly due to overlapping anatomical redundancy with the NDI image.

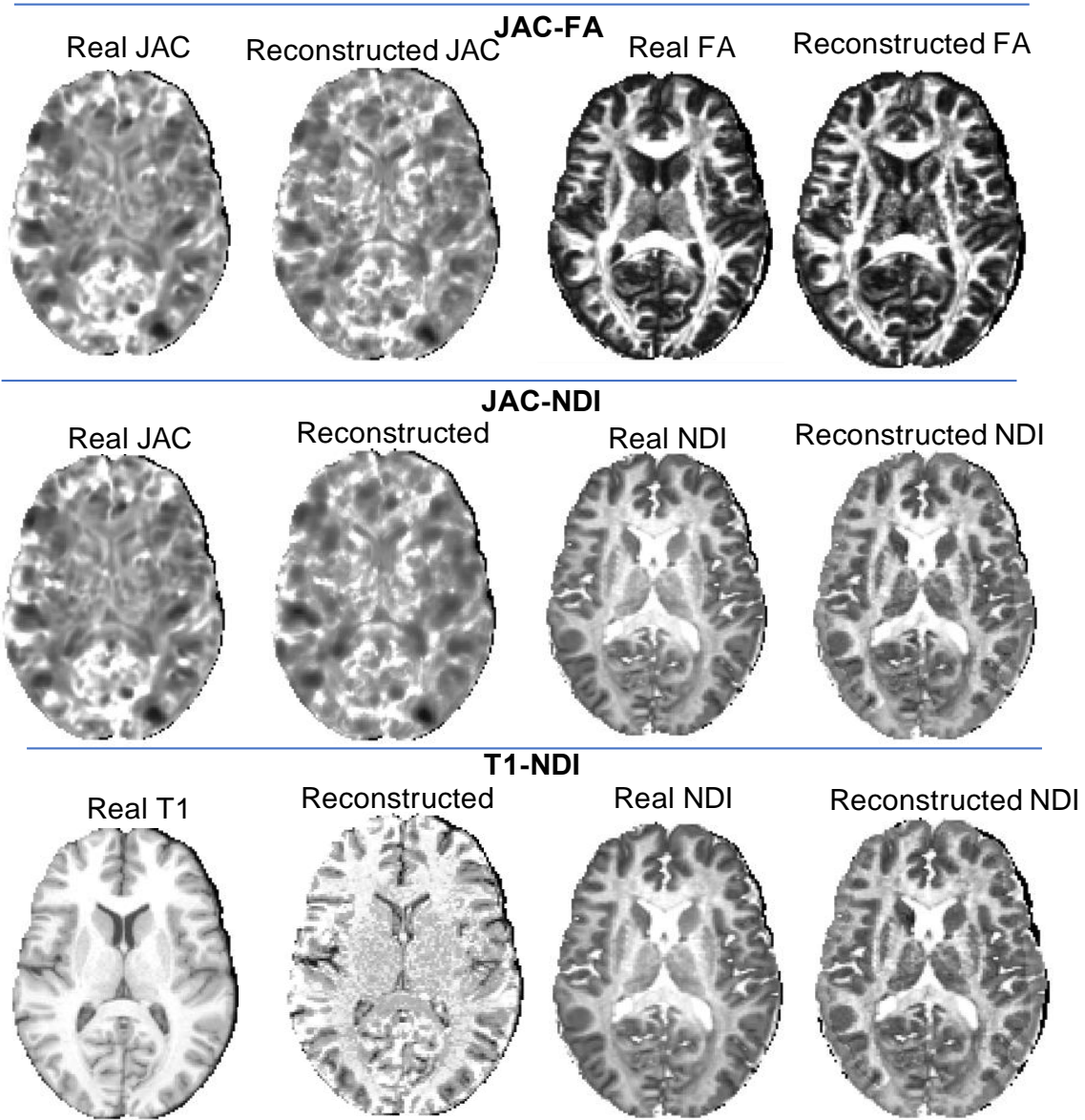

*Figure 4: Example of pairs of decoding for JAC-NDI, JAC-FA and T1-NDI modalities, using our multimodal architecture. Colour-scales are arbitrary but equal for each real-decoded pair.*

### 3.2. Cross-validation results

In this section, we present the cross-validation results (i.e. the generalization ability of our model) for one exemplary pair of modalities, namely JAC and NDI. This choice was made based on the facts that 1) the Jacobian determinant of a transformation which brings a T1 image into standard space has been seen to be a superior indicator of grey matter density changes as opposed to the T1 image intensities alone; 2) the NDI has shown promise and applicability in a vast number of neuroscience studies (along with a greater sensitivity and specificity in detecting microstructural alterations). Figure 5 depicts the 10-fold cross-validation results as described in the Methods section. MSE was bounded between 0 and 0.02 in all folds, except for fold six where it reached a median value (across all subjects in the fold of approximately 0.038). Considering that both JD and NDI metrics are bounded by definition between -1,1 and 0,1 (respectively) and commonly assume absolute values up to 0.5-0.7, these MSE values are around $1/200^{th}$ of the original intensities.

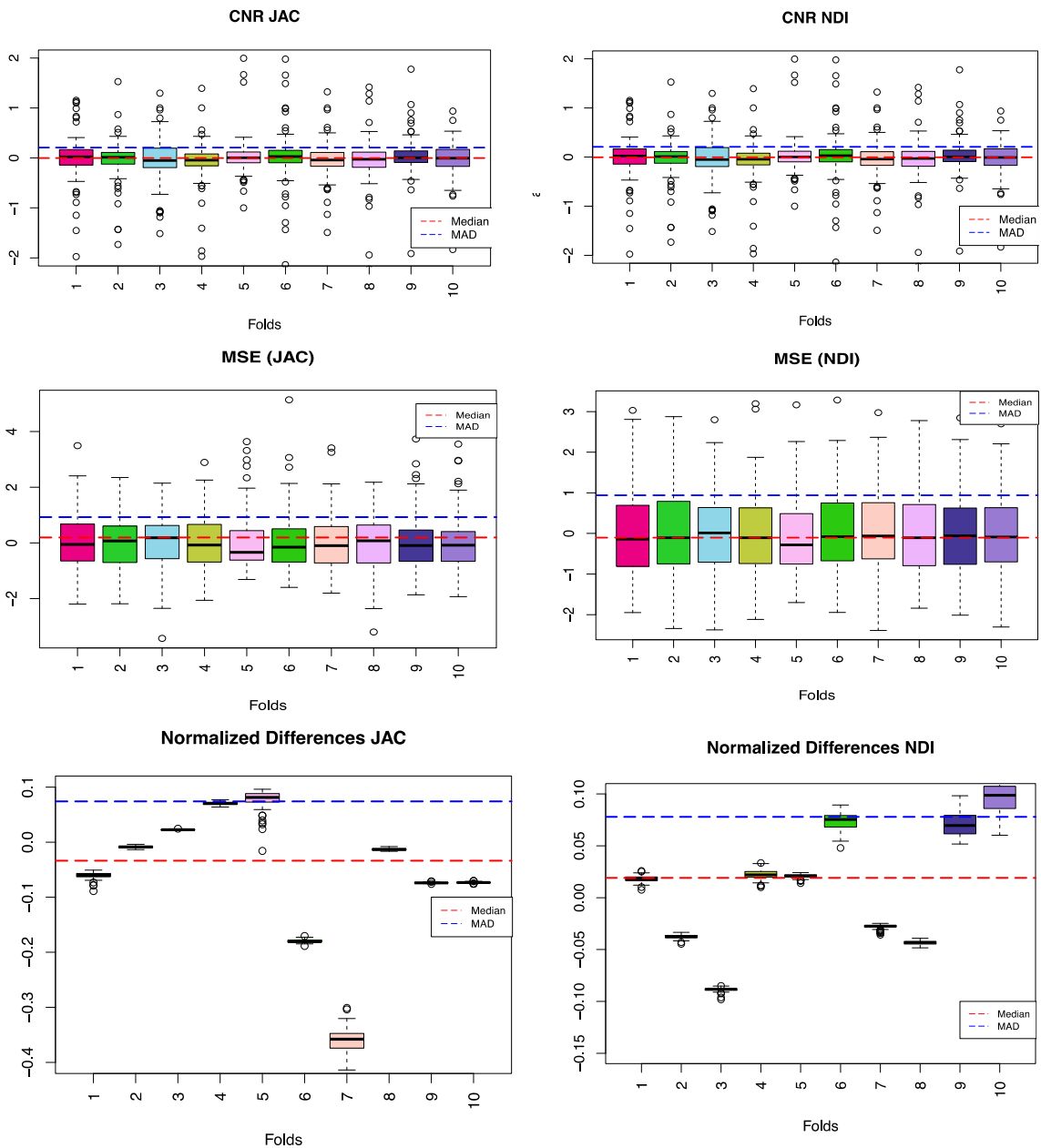

*Figure 5: CNR, MSE and Normalized Differences for each of the 10 folds within the cross-validation procedure. The boxplots represent the distribution of the values, where the median and the interquantile ranges of the distribution are plotted. MSE is plotted using the scaled absolute logarithm value of the original MSE. We also report the values of the Median (with the dotted red line) and of the Median Absolute Deviation (MAD, blue dotted line) across all folds. The x axis represents the folds (from 1 to 10) and the y axis the values of the respective plotted indicator.*

Running systematic experiments across combination of input and decoded modalities/contrasts in a 70/30 train/test split fashion, yielded the results shown in Table 2. MSE results were generally good (confined to a few % or less) and stable across different combinations. CNR results followed the same trend, with the exception of the second column from the left. While more extensive experiments would be needed to fully characterize this particular result, it is likely that it is related to the sampling nature of the definition we have created for the CNR metric.

| MSE Between True and Decoded Images | | | | |
|---|---|---|---|---|
| **Multi-modal inputs** | | | | |
| **JAC + NDI + FA +MD** | **JAC + NDI + FA** | **JAC + NDI** | **NDI** | |
| 0.01174 ± 0.000358 | 0.01915 ± 0.00123 | 0.00144 ± 3.55 x 10$^{-5}$ | 0.0004475 ± 3.58 x 10$^{-5}$ | **NDI** |
| 0.00248 ± 0.0002 | 0.02242 ± 0.00219 | 0.00044 ± 4.63 x 10$^{-5}$ | NA | **JAC** |
| 0.0007 ± 1.968 x 10$^{-5}$ | 0.1006 ± 0.0024 | NA | NA | **FA** |
| 0.0006 ± 4.73 x 10$^{-5}$ | NA | NA | NA | **MD** |
| Normalized Difference Between median CNR in True and Decoded Images | | | | |
| **Multi-modal inputs** | | | | |
| **JAC + NDI + FA +MD** | **JAC + NDI + FA** | **JAC + NDI** | **NDI** | |
| -0.0145 ± 0.011 | -0.2 ± 0.018 | 0.028 ± 0.009 | -0.017 ± 0.009 | **NDI** |
| 0.026 ± 0.01 | -0.266 ± 0.032 | -0.041 ± 0.009 | NA | **JAC** |
| -0.009 ± 0.009 | -0.249 ± 0.016 | NA | NA | **FA** |
| -0.01 ± 0.009 | NA | NA | NA | **MD** |

**Table 2.** Comparison results between true and decoded images (70/30 train/test split) with various modalities combinations. Columns represent the multi modal inputs, and the rows – the modality for which the metric is calculated (NA meaning measure not available as not present among the input modalities). The real images have been normalised in the 0-1 range. Results are given in the mean ± std format. "NA" refers to cases where a specific modality cannot be reconstructed since it is not part of the model inputs.

### 3.3. Phenotypic Variables and Factor Analysis

The independent phenotypical dimensions derived from the PCA-based factor analysis (see Methods) resulted in retaining 13 factors which cumulatively explained 63% of the total variance (Figure 6). Upon inspection of the factor loadings by an expert neuroscientist (LP), we were able to assign proof-of-concept, coherent interpretations to each factor based on the single phenotypic variables with highest loadings in each factor. In particular, in order of decreasing variance, these factors appeared to reflect negative affect; sociability; self-efficacy and wellbeing; crystallized intelligence; aggressive behavior; sustained attention and episodic memory; gender; fluid intelligence; high levels of blood pressure (hypertension); physical health/young age; positive personality traits (high agreeableness, high extraversion, and high mind openness); gait speed and physical endurance; right handedness.

| | Negative Affect | Sociality | Self-efficacy & Well-Being | Crystal IQ | Aggression | Memory Attention | Sex | Fluid IQ | BP | Young Age | E/O traits | Physical | Right Hand |
|---|---|---|---|---|---|---|---|---|---|---|---|---|---|
| Age | | | | | | | | | | -0.6 | | 0.4 | |
| Gender | | | | | | | 0.8 | | | | | | |
| AngAffect_Unadj | 0.8 | | | | | | | | | | | | |
| AngAggr_Unadj | | | | | 0.5 | | | | | | | | |
| AngHostil_Unadj | 0.5 | -0.3 | | | | | | | | | | | |
| ASR_Aggr_Raw | 0.5 | | | | 0.6 | | | | | | | | |
| ASR_Anxd_Raw | 0.7 | | -0.5 | | | | | | | | | | |
| ASR_Intr_Raw | | | | | 0.7 | | | | | | | | |
| ASR_Rule_Raw | | | | | 0.7 | | | | | | | | |
| ASR_Soma_Raw | 0.6 | | | | | | | | | | | | |
| ASR_Witd_Raw | 0.4 | -0.4 | | | 0.3 | | | | | | | | -0.3 |
| BMI | | | | | | | | | | -0.7 | | | |
| BPDiastolic | | | | | | | | | 0.9 | | | | |
| BPSystolic | | | | | | | | | 0.9 | | | | |
| CardSort_Unadj | | | | | | | | 0.7 | | | | | |
| DDisc_AUC_40K | | | | 0.3 | | | | | | | | | |
| Dexterity_Unadj | | | | | | | -0.4 | | | 0.3 | | | |
| EmotSupp_Unadj | | 0.7 | 0.3 | | | | | | | | | | |
| Endurance_Unadj | | | | | | | 0.3 | | | 0.5 | | 0.4 | |
| FearAffect_Unadj | 0.8 | | | | | | | | | | | | |
| FearSomat_Unadj | 0.6 | | | | | | | | | | | | |
| Flanker_Unadj | | | | | | | | 0.8 | | | | | |
| Friendship_Unadj | | 0.7 | | | | | | | | | | | |
| GaitSpeed_Comp | | | | | | | | | | | | 0.8 | |
| Handedness | | | | | | | | | | | | | 0.6 |
| InstruSupp_Unadj | | 0.6 | | | | | | | | | | | |
| IWRD_TOT | | | | | | 0.7 | | | | | | | |
| LifeSatisf_Unadj | | 0.3 | 0.6 | | | | | | | | | | |
| ListSort_Unadj | | | | 0.5 | | | | | | | | | |
| Loneliness_Unadj | 0.6 | -0.5 | | | | | | | | | | | |
| MeanPurp_Unadj | | 0.3 | 0.7 | | | | | | | | | | |
| NEOFAC_A | | | | | -0.6 | | | | | | 0.4 | | |
| NEOFAC_C | | | 0.5 | | -0.3 | | | | | | | | |
| NEOFAC_E | | 0.4 | 0.4 | | | | | | | | 0.5 | | |
| NEOFAC_N | 0.6 | | -0.5 | | | | | | | | | | |
| NEOFAC_O | | | | | 0.4 | | | | | | 0.6 | | -0.3 |
| PercHostil_Unadj | 0.5 | -0.4 | | | | | | | | | | | |
| PercReject_Unadj | 0.5 | -0.6 | | | | | | | | | | | |
| PercStress_Unadj | 0.7 | | -0.4 | | | | | | | | | | |
| PicSeq_Unadj | | | | | | | -0.3 | | | | | | |
| PicVocab_Unadj | | | | 0.8 | | | | | | | | | |
| PosAffect_Unadj | -0.3 | 0.3 | 0.6 | | | | | | | | | | |
| ProcSpeed_Unadj | | | | | | | | 0.7 | | | | | |
| ReadEng_Unadj | | | | 0.8 | | | | | | | | | |
| Sadness_Unadj | 0.8 | | | | | | | | | | | | |
| SCPT_SEN | | | | | | 0.8 | | | | | | | |
| SCPT_SPEC | | | | | | 0.8 | | | | | | | |
| SelfEff_Unadj | | | 0.6 | | | | | | | | | | |
| SSAGA_Educ | | | | 0.6 | | | | | | | | | |
| Strength_Unadj | | | | | | | 0.8 | | | | | | |
| VSPLOT_TC | | | | 0.4 | | 0.3 | | | | | | | |

*Figure 6. Loadings of each phenotypic variables (thresholded at an absolute value of 0.3 for visualization purposes) on each of the 13 factors which, cumulatively, explained 63% of the data variance. For a detailed explanation of the phenotypic variables and their origin, please see the HCP data dictionary at https://wiki.humanconnectome.org/display/PublicData/HCP+Data+Dictionary+Public-+Updated+for+the+1200+Subject+Release. See supplementary materials for descriptive statistics of the original phenotypic variables.*

## 3.4 Clustering and mapping multimodal embeddings to phenotypic variables

When applying the clustering procedure to the multimodal embeddings computed (as described above) from the JAC and NDI images, we obtained nine distinct clusters (see Fig. 7 for sizes and for colour reference for subsequent results on phenotypic variables). Figure 7 shows the number of individuals belonging to each cluster for the JAC-NDI experiments performed, as well as a low-dimensional representation of the high-dimensional embeddings which visually confirms the existence and separations of the clusters we found.

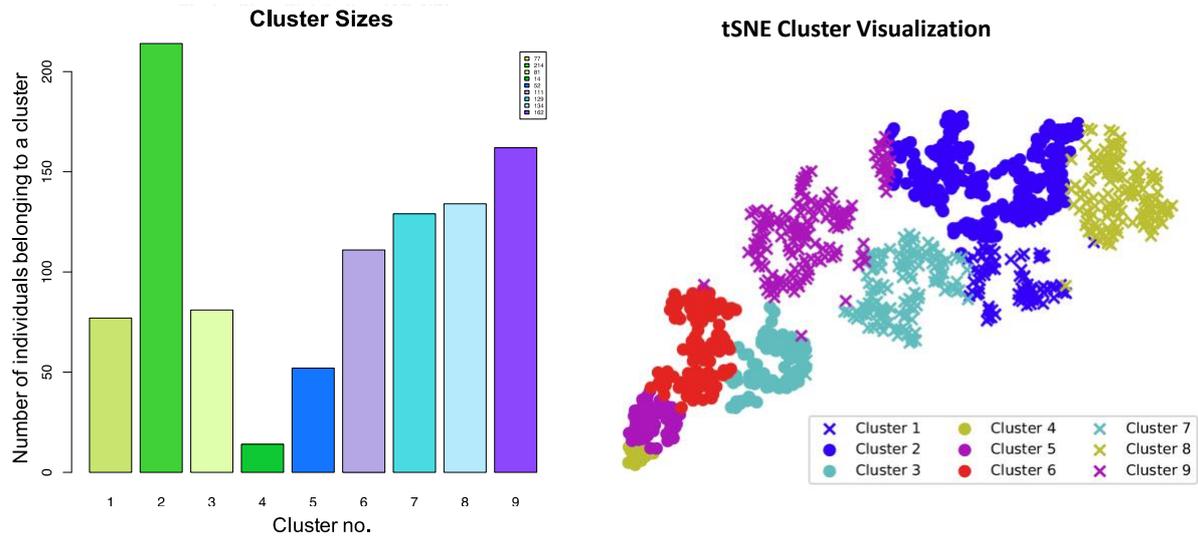

*Figure 7: Left: Clusters sizes obtained when running Affinity Propagation clustering with optimized parameters (see Methods) on the multimodal embeddings for the JAC-NDI embeddings. Right: tSNE visualization of the high-dimensional embeddings, color-coded according to clusters found by affinity propagation.*

When using the KW test to explore between-cluster differences in factors extracted from the phenotypic variables, we obtained significant effects in six factors which had been identified as representing (in this sample) aggression, Blood Pressure, Crystallized Intelligence, Righthandedness, biological sex and Young age (FDR corrected across all factors), see Fig 8. In addition, for all these values, bootstrap analysis (see methods) showed that these effects (which were due to the assignment of subjects to each cluster of a certain size) were far not random (see Fig 9.).

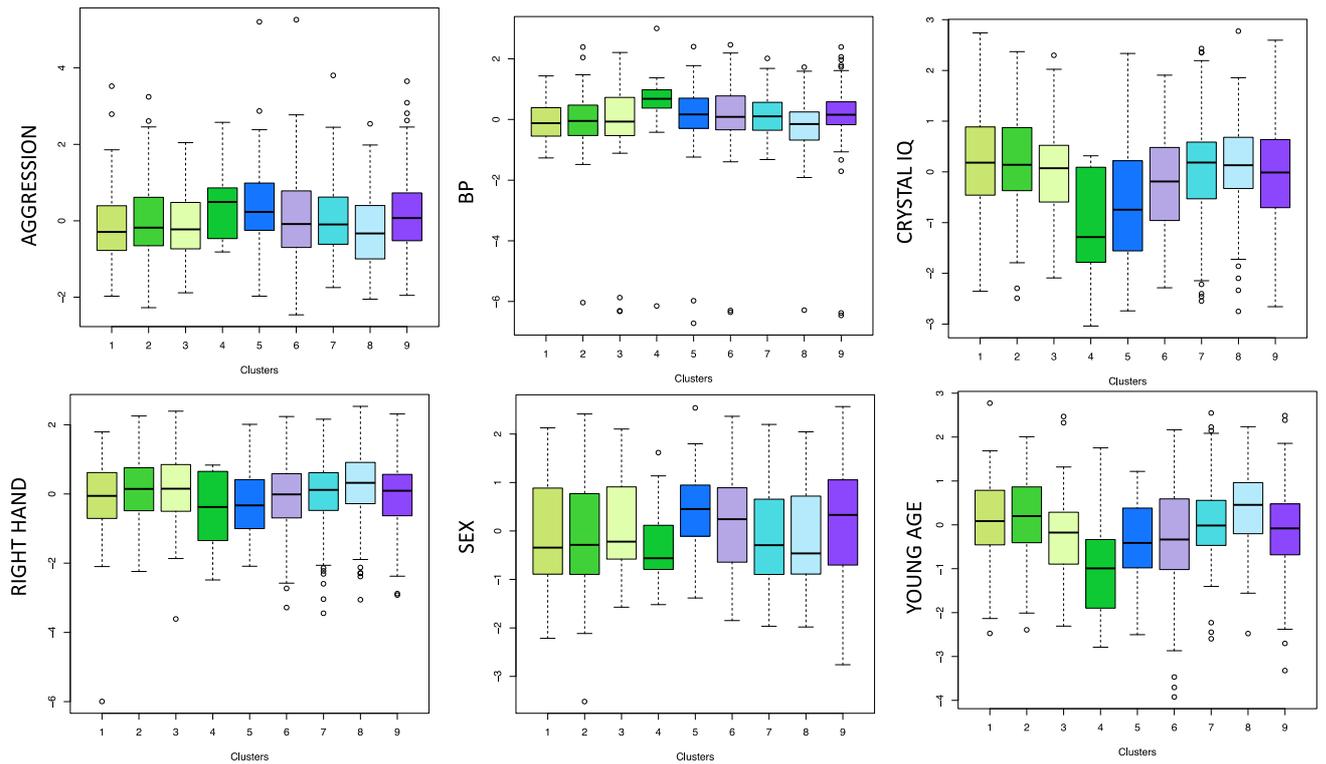

*Figure 8: Phenotypic factor distribution across the nine identified clusters. Only phenotypic factors in which a significant difference (FDR corrected across clusters) across clusters are shown. Same colour codes for clusters as per Figure 7 are used.*

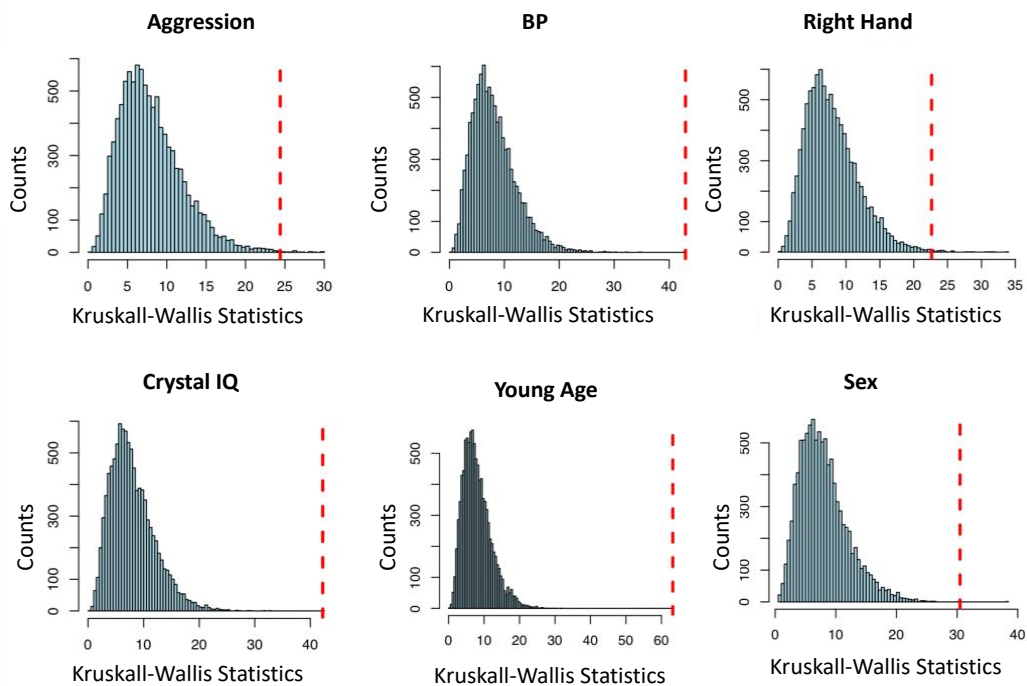

*Figure 9: Bootstrap results comparing the "true" KW statistics (red, dotted line) resulting from our cluster assignments to the distribution of the KW statistics obtained when comparing Factor values between clusters after repeatedly (10000 samples) and randomly sampling individuals across cluster distributions with the same sizes.*

Finally, we asked the question whether the clusters we identified corresponded to recognizable/interpretable "profiles" (made of phenotypic factors) which could be further interpreted and/or heuristically explained. To this end, for each cluster and each factor, we computed the cluster-wise median of each factor and evaluated its position (quantile) in the whole population. Figure 10 summarizes these results. It is important to note that is not a clustering result per se (i.e. the hierarchical clustering brackets on the left are only included for visualization purposes). Rather, the figure represents an overall impression of how the median value of each factor and each cluster is positioned with respect to the whole population. This allows visual inspection of the overall idiosyncrasy of each cluster as a whole, with respect to the whole population as well as with respect to other clusters.

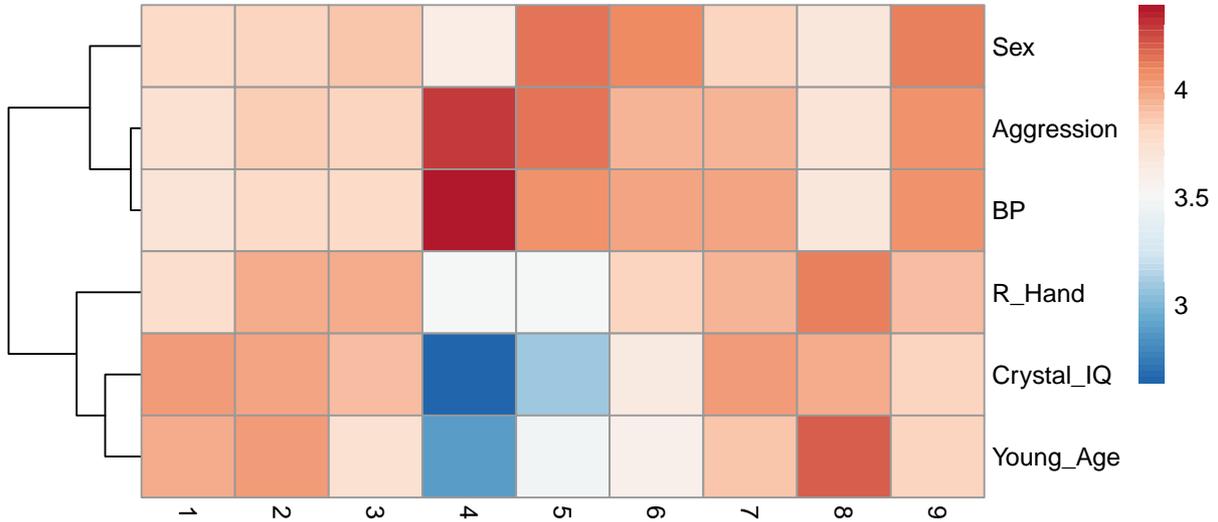

*Figure 10: Cluster "profiles" Position (logarithm of the quantile of cluster-wise median of each factor with respect to the whole population distribution). Columns therefore represent cluster "profiles" as characterized by comparing their median value in each phenotypic factor to the distribution of the same factor across the population.*

Interestingly, it is fairly evident how unsupervised clustering of our embeddings, generated from multimodal brain imaging only, not only yields separations which are non-random with respect to permuting individuals, but also that the clusters created have, to some extent, combinations of values of phenotypic traits which appear to delineate distinct, biologically consistent groups of individuals. An example would be cluster 4 composed by people with high blood pressure, high aggression, below-average crystal IQ and older age. While such an extreme cluster in hard to detect in a non-clinical sample of health people, it was picked up by our approach in spite of its small dimension (14 people). Cluster 5 appears to go in the same direction (albeit in a less extreme way), while being more polarized towards female sex.

## Discussion

One of the main goals of this paper was to handle the large amounts of information present in not only one, but multiple medical imaging modalities in a large number of subjects, maximizing information content and "compression" at the same time. Our framework is based on a modular approach and separable convolutional blocks, which can efficiently deal with highly dimension datasets. In fact, taking a conventional 3D convolutional block as baseline, 3D conventional convolution requires d1xd2xd3xCxC' parameters, whereas a 3D separable convolutions requires d1xd2xd3 + CxC' parameters. The image spatial dimensions are given by d1, d2, d3, and the number of input and output channels by C and C' respectively. Therefore, 3D separable convolutions use $\frac{1}{C'} + \frac{1}{d1xd2xd3}$ times fewer parameters. Using the kernel sizes and number of channels for our architecture (Fig. 2 and Fig. 3), this translates into $\frac{1}{36} + \frac{1}{96} = 0.038$, or ~26 times fewer parameters for the 3D separable variant. Given that 3 out of 5 convolutions are separable depthwise convolutions in each step of our architecture (the remaining 2 are conventional), this means that our architecture comprises only ~40% parameters of its fully conventional convolutional equivalent, hence reducing GPU memory utilization by 60%.

We aimed for the architecture to be generative in order to directly evaluate the decode part in terms of various decoding metrics. This ensures in turn that out low dimensional embeddings 1) succeed in encoding meaningful information, and 2) are able to regenerate the two modalities separately in spite of the complete information fusion implemented in the last encode step. Further, our deep learning architecture, which is heavily based on parameter-efficient separable convolutional blocks, is very lightweight, which reduces both training time and GPU memory utilization, therefore making our framework practical to handle multiple high-resolution 3D medical image modalities. In spite of the large dimensionality reduction, we obtained good decoding quality in MSE, normalized differences and also CNR, an additional metric commonly used in evaluating the usefulness of medical images. In comparison with more traditional methods like e.g. PCA, it important to point out that conventional PCA is equivalent to an autoencoder with a single fully connected hidden layer (i.e. the produced embeddings would span the same subspace in both cases (P. Baldi and K. Hornik, 1989; H. Bourlard and Y. Kamp. 1988). Consequently, an autoencoder with a sequence of non-linear hidden layers is equivalent to non-linear PCA. In this sense, our deep autoencoder can be viewed as non-linear PCA, and the embedding space it generates spans the space of conventional PCA as a possible solution in addition to other solutions. Also, given that PCA requires singular value decomposition of the input matrix, and that a dataset of ~1000 subjects, several modalities per subject, and up to $10^6$ voxels for each modality, a direct PCA would be computationally prohibitive.

One notable prior study has presented an integrative analysis of features extracted from brain structural MRI, finding evidence for a subspace which is able to explain a large share of the inter-individual variation of brain structure in healthy people (Llera et al 2019). The authors also identified relationships between inter-individual brain *structural* variations and numerous demographic and behavioural variates, in particular demonstrating that a spectrum of behavioural and demographic data variability can be identified and explained using structural features only. However, the feature extraction method used in this paper requires a-priori information about which features to extract, hence strongly limiting generalizability. Other application of DL for dimensionality reduction in the field of medical imaging exist, see e.g. Benson et al., 2014 or Lundervold et al., 2019 for a review. However, most of these papers rely on a priori feature engineering like e.g. aggregating information by averaging within ROIs, hence heavily reducing the dimension of the problems fed to the actual DL architecture. Other methods rely on the extraction of patches (e.g. Saha et al., 2020, Kao et al., 2020) of a priori segmentation of white and grey matter volumes. Our work, however, differs in that we use image decoding simply to learn lower dimensional embeddings of multi-modal MRI data, find structure in the representations within this space, and demonstrate in a specific use-case that this structure can, for instance, correspond to interpretable sub-types. On the other hand, these latter recent papers focus on improving the quality of image synthesis and address the issue of data scarcity by improving learning efficiency with no focus on dimensionality reduction and the discovery of latent factors. A key issue here is that the question whether the information included in our high-dimensional embeddings is "necessary or important" is strongly use-case dependent. Since the method is unsupervised the "importance" of the captured information can be estimated by the plausibility and contextualization of the results obtained in whatever downstream application will be performed using the latent embeddings.

In our proof-of-concept application, we focused on an example pair of modalities, JAC and NDI. The Jacobian Determinant, or JD, is a voxel-wise measure of the volumetric changes that are computed from the deformation needed to precisely match each subject-specific structural MR image to a population-specific MRI template. Relative to the more commonly used voxel based morphometry (VBM), tensor-based morphometry analyses derive JDs volumetric changes from non-linear warp fields, rather than from signal intensities alone (Avants et al., 2011; Chung et al., 2001; Villalon et al., 2011). In turn, the NDI has been associated with microscopic variability in the neurite architecture and geometry, highlighting the viability of the NDI as a proxy measure for biological changes in the neuropil (i.e. dendrites, axons, and glial filaments). The NDI is also considered a more sensitive diffusion-imaging metric compared to the canonical diffusion tensor imaging (DTI) imaging measures such as FA, MD, and RD. Notably, when clustering the embeddings in an unsupervised manner, we were able to demonstrate the creation of clusters based on brain data only, which map to differences in phenotypic dimensions. In addition to checking statistical significance of those differences, we also demonstrated through bootstrap analysis that those differences in phenotypic dimensions are not random with respect to the partitioning generated by the clustering algorithm. In addition, we also demonstrated robustness and decode fidelity across a wider range of input and decode combinations (Table 2). As an example, when decoding NDI from 1,2,3, or 4 modalities, the MSE remains below 1% maximum, while reconstructing JAC the MSE remains bounded within 2% maximum. In this context, it should be noted that while in principle a fusion model with a higher number of modalities would surpass a model with fewer ones, this statement would need to be modulated in terms of model capacity, data availability and, importantly, use-case. There may therefore be different optimal number of modalities for different use-cases, which we feel should be investigated in future work based on the architecture we present here. Also, when examining the phenotypic "profiles" of these clusters, in spite of the narrow age range (25-45) and, in general, great homogeneity of the HCP sample of healthy individuals (i.e. comparatively little variance to encode), we were able to detect both some expected and sensible associations, and some interesting, and novel characteristics of the cohort we stratified.

For example, cluster 4 grouped phenotypically 'extreme' male individuals with high levels of aggression, high blood pressure, relatively older age and lower crystallized IQ. Likewise, cluster 5 contains female participants displaying similar demographic, health, and behavioural features as those in cluster 4 (particularly the association between aggression and low IQ). Although the number of individuals grouped within each of these empirically derived categories is relatively low (n=14 for cluster 4 and n=52 for cluster 5), the association between aggression and low IQ is interesting and expected on the basis of previous literature. More specifically, it highlights the proof-of-concept ability of our architecture to detect relevant phenotypic associations in non-clinical samples even when they exist only in a small subset of subjects. We feel that, in addition to heuristically validating the information content of our multimodal embeddings, these results demonstrate the possibility of a brain-to-phenotype mapping which can easily be extended to the important application of identifying pathological substrates for phenotypic disturbances. Relatedly, in (Llera et al 2018) the authors present a comprehensive analysis of features extracted from brain MRI, features, linking the inter-individual brain structure variability to a large set of behavioural and demographics scores. Again however, in this paper the features are extracted a-priori from the input dataset and are not learnt during training, rendering this method application-specific. Numerous examples in the literature demonstrate the usefulness of unsupervised approaches to stratifying patients such as cancer, or neurodegenerative pathologies as for example Alzheimer (Lopez et al 2018, Lasko et al 2013, Wang et al 2013, Wu et al 2017, Alashwal et al 2019, Young et al 2017). Similar studies allowed to predict patient evolution based on clusters generated from both cerebrospinal fluid MRI data (Nettiksimmons et al., 2010). Also, previous work clustered AD patients using, e.g., the CLustering In QUEst (CLIQUE) strategy (Gamberger et al., 2016a, 2016b), hierarchical clustering or k-means clustering for profiling AD patients, older adults in general (Escudero et al., 2012; Zemedikun et al., 2018), or patients with other disorders, including aphasia (Hoffman et al., 2017). Again, all these studies rely on some sort of prior engineering which pre-emptively eliminated the need for learning data reduction during training. Also, initial efforts have been made towards a biomarker-based characterization of substrata of neurological disease. For example, the so-called A/T/N categorization is a simple conceptual clinical and biological construct which may, to some extent, stratify the pathophysiological *continuum* of AD and other neurodegenerative disorders according to the presence of established and reproducible AD-related pathophysiological markers (Bennett et al., 2016). Still, from a machine learning perspective, in this type of procedure the feature engineering and learning steps are disjoint (hence sacrificing representational ability) and the generalizability of such categories as well as the optimality of the cut-offs employed in generating them has not been investigated. Unbiased stratification of the AD *continuum* was also explored using CSF markers alone (Toschi et al., 2019). Similar initial attempts have been made in PD (Zhang et al., 2019) where three robust subtypes of PD were identified. The existence of separate clusters in embedding space (which also separate in terms of specific phenotypic variables) is presented as an example use-case, and including different phenotypical variables may have yielded larger (or smaller) differences between clusters. Also, while the non-randomness of the differences between cluster medians in phenotypical variables, has been demonstrated through bootstrap experiments, the cohort employed in these experiments is not representative of community dwelling population. Instead, the HCP database is composed highly selected, healthy young adults, between 22 and 40 years old, which are therefore expected to exhibit less variance than a more representative age group and even less variance than, e.g. a pathological spectrum (like e.g Parkinson's or Alzheimer's disease.

Some caveats should be mentioned. For example, some of the image contrasts we included in our experiments (e.g. FA, NDI etc) are actually derived from the same imaging modality (DWI). While it would have been interesting to include e.g. raw DWI data in our experiments, we would have had to make strong *a priori* decisions about a handful of gradient directions to include out of the 500+ diffusion weighted volumes available for each subject. Also, in this paper we did not optimize hyperparameters. For example, varying embedding length would affect both reconstruction quality and the outcome of any specific use. The hyperparameters were chosen from our previous work (Spasov et al. 2019), which uses a deep learning architecture which relies on the same operational blocks. While the decode fidelity of our architecture could be further improved by hyperparameter optimization, it is important to note that this would require nested cross validation. Given that a single cross-validation experiment (i.e. without hyperparameter optimization) with only two modalities takes approximately 2 weeks to run on a Titan V GPU with approximately 5000 tensor cores, nested cross validation would be computationally intractable even with large cloud computing resources (which, incidentally, explains why several deep learning papers do not focus on hyperparameter optimization). Still, the current set of hyperparameters achieve very good performance when it comes to image reconstruction quality (MSE between reconstructed and ground truth images is approximately $1/200^{th}$ of the original intensities). Also, while it would be interesting to compare our results with other architectures, given that our workflow also includes clustering of the latent space, subsequent stratification a healthy subject population, and interpretation of these strata in terms of phenotypical variables, baselines which constitute a fair comparison would need to be partially engineered ad hoc. Also, if the clustering use-case of our embeddings was to be turned into a clinical application (e.g. assigning an unseen subject to a particular, previously identified cluster), external validation of the unsupervised clustering procedure would be appropriate. However, the aim of this example was to demonstrate that our latent embeddings contain information that is not only separable but also possibly useful for exploring whether brain-derived information can be mapped to exogenous variables at all. Also, this paper does not explicitly analyse model robustness to e.g. missing inputs or missing modalities. While the main goals of this study were to design the novel architecture and provide proof of concept of its usefulness, robustness to missing inputs at various level would be important for systems used 'in production' (e.g. that is within a clinical environment, after identification of a suitable task). Summarizing, potential improvements would be, for example, the inclusion of raw data experimentation, hyperparameter optimization, robustness check to missing inputs and finally the need of further experimentation to turn the method proposed in a clinical support system tool. We therefore believe these aspects will be all part of future work experimentation and testing.

In summary, we created a deep learning architecture based on parameter-efficient building blocks like e.g. separable convolutions (which result in a 20-fold decrease in parameter utilization in a single mid-flow block) which is able to compress multiple voxel-wise brain images with minimal loss of information, good generalization abilities (in cross validation) and whose latent embeddings appear to map to meaningful and interesting phenotypic profiles. In view of possible clinical applications, the predictive validity of our clusters would need to be corroborated by longitudinal follow-ups in order to test whether the 'extreme' groups of people identified go on and develop clinically relevant disorders. If this is the case, this would strongly reinforce the idea that deep learning information compression based exclusively on neuroimaging has the ability to predict clinically relevant behavioural trajectories. We expect that our model will be able to aid in the current quest for solid avenues towards personalized medicine, i.e. the goal of creating an individual patient profile which is matched exactly as possible to a diagnosis, intervention and prognosis. Detecting generalizable patient strata from brain images only can aid in creating multi-dimensional biomarkers able to chart spatio-temporal trajectories of (possibly) identifiable pathophysiological mechanisms. It is hoped that such patients (sub)strata, identified through fully data-driven approaches, will correspond to e.g. different genetic makeups, different phenotypic groups, end eventually will respond differently to specific treatment strategies. If this were true, such results may be of aid in predicting disease evolution as well as drug response, hence also empowering clinical trials. In turn, this would provide stepping stones for the development of e.g. pharmacological treatments tailored to biomarker-guided homogeneous subgroups (or clusters).

## Acknowledgements

Simeon Spasov research is supported by EPSRC. Luca Passamonti is funded by the Medical Research Council (MRC) grant (MR/P01271X/1) at the University of Cambridge, UK. The GPUs on which this work was performed were generously provided by NVIDIA. Part of this work has been supported by the European Union's Horizon 2020 research and innovation programme under grant agreement No. 101017727 (EXPERIENCE project).